%% file: main.tex
\documentclass[
]{ceurart}

\usepackage[linesnumbered,ruled,vlined]{algorithm2e} 
\usepackage{listings}
\usepackage[most]{tcolorbox}
\usepackage{caption}
\lstdefinelanguage{json}{
    basicstyle=\ttfamily\small,
    showstringspaces=false,
    breaklines=true,
    string=[s]{"}{"},
    morestring=[b]',
    literate=
     *{0}{{{\color{blue}0}}}{1}
      {1}{{{\color{blue}1}}}{1}
      {2}{{{\color{blue}2}}}{1}
      {3}{{{\color{blue}3}}}{1}
      {4}{{{\color{blue}4}}}{1}
      {5}{{{\color{blue}5}}}{1}
      {6}{{{\color{blue}6}}}{1}
      {7}{{{\color{blue}7}}}{1}
      {8}{{{\color{blue}8}}}{1}
      {9}{{{\color{blue}9}}}{1}
      {:}{{{\color{black}:}}}{1}
      {,}{{{\color{black},}}}{1}
      {"}{{{\color{red}"}}}{1},
}

\newtcblisting{prettyjson}{
    listing only,
    listing options={
        language=json,
        basicstyle=\ttfamily\footnotesize,
        breaklines=true,
        columns=fullflexible,
        frame=none,
        showstringspaces=false,
        tabsize=2
    },
    colback=gray!3,
    colframe=gray!30,
    arc=8pt,
    outer arc=8pt,
    boxrule=0.3pt,
    left=4pt,
    right=4pt,
    top=2pt,
    bottom=2pt,
    enhanced,
}

\usepackage{newunicodechar}

\newunicodechar{¬}{\ensuremath{\lnot}}
\newunicodechar{∧}{\ensuremath{\land}}
\newunicodechar{∨}{\ensuremath{\lor}}
\newunicodechar{∃}{\ensuremath{\exists}}
\usepackage{listings}
\usepackage{pifont}
\renewcommand{\checkmark}{\ding{51}}%
\newcommand{\xmark}{\ding{55}}%
\newcolumntype{M}[1]{>{\centering\arraybackslash}m{#1}}
\newcolumntype{L}[1]{>{\raggedright\arraybackslash}m{#1}}
\usepackage{hyperref}

\begin{document}

\copyrightyear{2025}
\copyrightclause{Copyright for this paper by its authors.
  Use permitted under Creative Commons License Attribution 4.0
  International (CC BY 4.0).}

\conference{ITADATA2025: The 4$^{\text{th}}$ Italian Conference on Big Data and Data Science, September 9--11, 2025, Turin, Italy}

\title{Bridging LLMs and Symbolic Reasoning in Educational QA Systems: Insights from the XAI Challenge at IJCNN 2025}


\author[1]{Long S. T. Nguyen}[%
orcid=0009-0008-7488-4714,
email=long.nguyencse2023@hcmut.edu.vn,
]
\fnmark[1]
\address[1]{URA Research Group, Ho Chi Minh City University of Technology (HCMUT), Vietnam}

\author[1]{Khang H. N. Vo}[%
orcid=0009-0009-7631-2450,
email=khang.vo872003@hcmut.edu.vn,
]
\fnmark[1]

\author[1]{Thu H. A. Nguyen}[%
email=thu.nguyenhoanganh14@hcmut.edu.vn,
]

\author[1]{Tuan C. Bui}[%
email=tuanbc88@hcmut.edu.vn,
]

\author[1]{Duc Q. Nguyen}[%
email=nqduc@hcmut.edu.vn,
]

\author[2]{Thanh-Tung Tran}[%
email=tttung@hcmiu.edu.vn,
]
\address[2]{Ho Chi Minh City International University (HCMIU), Vietnam}

\author[3]{Anh D. Nguyen}[%
email=anh.nguyen.duc@usn.no,
]
\address[3]{University of South-Eastern Norway, Norway}

\author[4]{Minh L. Nguyen}[%
email=nguyenml@jaist.ac.jp,
]
\address[4]{Japan Advanced Institute of Science and Technology (JAIST), Japan}

\author[5]{Fabien Baldacci}[%
email=fabien.baldacci@labri.fr,
]
\address[5]{Univ. Bordeaux, CNRS, Bordeaux INP, LaBRI, UMR 5800, F-33400 Talence, France}

\author[1]{Thang H. Bui}[%
email=bhthang@hcmut.edu.vn,
]

\author[6]{Emanuel Di Nardo}[%
email=emanuel.dinardo@uniparthenope.it,
]
\address[6]{University of Naples Parthenope, Italy}

\author[6]{Angelo Ciaramella}[%
email=angelo.ciaramella@uniparthenope.it,
]

\author[7]{Son H. Le}[%
email=sonlh@vnu.edu.vn,
]
\address[7]{VNU Information Technology Institute, Vietnam National University, Vietnam}

\author[8]{Ihsan Ullah}[%
email=ihsan.ullah@universityofgalway.ie,
]
\address[8]{Visual Intelligence Lab, School of Computer Science \& Insight Center for Data Analyitcs, University of Galway, Ireland}

\author[9]{Lorenzo Di Rocco}[%
email=lorenzo.dirocco@uniroma1.it,
]
\address[9]{Sapienza University of Rome, Italy}

\author[1]{Tho T. Quan}[%
orcid=0000-0003-0467-6254,
email=qttho@hcmut.edu.vn,
]
\cormark[1]

\cortext[1]{Corresponding author.}
\fntext[1]{These authors contributed equally.}

\begin{abstract}
The growing integration of Artificial Intelligence (AI) into education has intensified the need for transparency and interpretability. While hackathons have long served as agile environments for rapid AI prototyping, few have directly addressed eXplainable AI (XAI) in real-world educational contexts. This paper presents a comprehensive analysis of the XAI Challenge 2025, a hackathon-style competition jointly organized by Ho Chi Minh City University of Technology (HCMUT) and the International Workshop on Trustworthiness and Reliability in Neurosymbolic AI (TRNS-AI), held as part of the International Joint Conference on Neural Networks (IJCNN 2025). The challenge tasked participants with building Question-Answering (QA) systems capable of answering student queries about university policies while generating clear, logic-based natural language explanations. To promote transparency and trustworthiness, solutions were required to use lightweight Large Language Models (LLMs) or hybrid LLM–symbolic systems. A high-quality dataset was provided, constructed via logic-based templates with Z3 validation and refined through expert student review to ensure alignment with real-world academic scenarios. We describe the challenge’s motivation, structure, dataset construction, and evaluation protocol. Situating the competition within the broader evolution of AI hackathons, we argue that it represents a novel effort to bridge LLMs and symbolic reasoning in service of explainability. Our findings offer actionable insights for future XAI-centered educational systems and competitive research initiatives.

\end{abstract}


\maketitle

\section{Introduction}
\input{components/Introduction}

\section{Positioning Within the Landscape of AI Competitions}

\input{components/Related_Works}

\section{The XAI Challenge 2025}

\subsection{Motivation and Objectives}
\input{components/Objectives}

\subsection{Structure and Timeline}
\input{components/Event_Structure}

\subsection{Dataset} \label{data}
\input{components/Dataset}

\subsection{Rules and Constraints}
\input{components/Rules}

\subsection{Evaluation Protocol \label{sec:evaluation}}
\input{components/Evaluation}

\subsection{Participant Overview}
\input{components/Participants}

\subsection{Results and Analysis}
\input{components/Results}

\section{Selected Approaches}
\input{components/Solution}

\section{Conclusion}
\input{components/Conclusion}

\begin{acknowledgments}
We would like to express our sincere gratitude to the URA Research Group at Ho Chi Minh City University of Technology (HCMUT), Vietnam, especially the undergraduate students who contributed to building and reviewing the dataset. We also acknowledge Bao Gia Quach, Hung Canh Nguyen, Nguyen Bao Le, Hieu Tran Hoang Nguyen, Hoang Huy Vu, Thuong Tran Anh Le, and Quynh Thi Nhu Vo for their support in both technical coordination and team communication throughout the challenge.

Finally, we are grateful to Professor Akka Zemmari and Professor Pascal Desbarats from the University of Bordeaux, France, for their visit to HCMUT and valuable discussions during the early proposal phase, together with Professor Fabien Baldacci, co-author of this paper.
\end{acknowledgments}

\section*{Declaration on Generative AI}

  The author(s) have not employed any Generative AI tools.
  

\bibliography{ref}  

\end{document}

%% file: components/Introduction.tex
Hackathons emerged in the late 1990s as intensive, time-constrained events where developers collaborated to rapidly prototype functional solutions \cite{briscoe2014digital}. Initially focused on general-purpose programming, these events gradually evolved into innovation incubators across diverse domains. By the early 2010s, hackathons had become increasingly integrated into educational contexts, offering informal yet impactful environments for learners to connect theoretical understanding with real-world application \cite{porras2019code}. They fostered creativity, collaboration, and technical fluency, which are essential skills in the rapidly advancing field of \textit{Artificial Intelligence} (AI). The mid-2010s marked a turning point, as breakthroughs in machine learning and the rise of open-source frameworks such as TensorFlow\footnote{\url{https://www.tensorflow.org/}} and PyTorch\footnote{\url{https://pytorch.org/}} enabled hackathons to address more complex and data-driven problems \cite{komssi2014hackathons}. In the early 2020s, the emergence of \textit{Large Language Models} (LLMs), including ChatGPT\footnote{\url{https://openai.com/index/chatgpt/}} and GitHub Copilot\footnote{\url{https://github.com/features/copilot}}, significantly enhanced participants’ productivity, code quality, and learning outcomes. At the same time, these tools raised concerns about academic integrity and over-reliance on automation, highlighting the need for clearer ethical guidelines and responsible deployment \cite{su15075614, bdcc8120188}.

As LLMs became increasingly capable but remained difficult to interpret, \textit{eXplainable AI} (XAI) emerged as a crucial paradigm for promoting transparency. This need is especially pronounced in educational settings, where students and educators often seek justifications for automated decisions \cite{8466590}. In parallel, symbolic AI has regained attention as a complementary approach to purely data-driven methods. A notable milestone was the 2024 release of AlphaGeometry by Google DeepMind\footnote{\url{https://deepmind.google/}}, a neuro-symbolic system that achieved gold medal-level performance at the International Mathematical Olympiad. At its core lies a symbolic reasoning engine, which demonstrates the potential of logic-based inference in addressing complex educational problems \cite{Trinh2024, GARNELO201917}. In this broader landscape, AI competitions have begun to incorporate educational and explainability-oriented goals. For example, EfficientQA \cite{pmlr-v133-min21a} and the Alexa Prize TaskBot\footnote{\url{https://www.amazon.science/alexa-prize/taskbot-challenge}} focused on answer accuracy and dialogue, but offered limited support for interpretable outputs. Symbolic reasoning benchmarks such as the \textit{Abstraction and Reasoning Challenge} (ARC)\footnote{\url{https://www.kaggle.com/competitions/abstraction-and-reasoning-challenge}} and Neuro-Symbolic ARC \cite{batorski2025nsa} emphasized structured reasoning but did not require \textit{natural language} (NL) explanations. Similarly, biomedical competitions such as the MIDRC mRALE Challenge\footnote{\url{https://www.midrc.org/xai-challenge-2024}} targeted explainability in medical diagnostics, rather than educational domains.

To fill this gap, the XAI Challenge 2025\footnote{\url{https://sites.google.com/view/trns-ai/challenge}} was introduced as part of the \textit{International Joint Conference on Neural Networks} (IJCNN 2025), co-organized by \textit{Ho Chi Minh City University of Technology} (HCMUT) and the \textit{International Workshop on Trustworthiness and Reliability in Neurosymbolic AI} (TRNS-AI). Although the challenge ran for three months, it was inspired by the spirit of hackathons, emphasizing rapid iteration, interdisciplinary collaboration, and practical relevance. The competition was structured into multiple phases, including dataset familiarization, system development, and explanation refinement. Participants were asked to build educational \textit{Question-Answering} (QA) systems that could respond to student queries about academic regulations while providing logically grounded natural language justifications. All solutions were required to use lightweight LLMs or hybrid LLM and symbolic reasoning models, with an emphasis on transparency, verifiability, and alignment with human logic. The challenge was supported by a carefully designed dataset grounded in real university policies. This dataset was developed through a two-stage pipeline: (i) synthetic examples were generated using logic-based templates and validated using the Z3 Solver \cite{10.1007/978-3-540-78800-3_24}, and (ii) these examples were refined and validated by trained university students to ensure clarity, factual accuracy, and educational relevance.

This paper presents a comprehensive overview of the XAI Challenge 2025, including its motivation, structure, dataset design, evaluation methodology, and participant outcomes. By situating the challenge within the broader evolution of AI competitions, we argue that it represents a new class of explainability-focused systems aligned with educational values. In particular, we emphasize how the challenge brings together the strengths of LLMs and symbolic reasoning, establishing a practical benchmark for trustworthy and pedagogically sound AI in education.

%% file: components/Related_Works.tex
In recent years, AI competitions have proliferated not only as benchmarks for measuring technical progress but also as collaborative platforms for exploring emerging paradigms such as LLMs, symbolic reasoning, and explainability. These events cover a wide range of tasks, including open-domain QA, task-oriented dialogue, program synthesis, and interpretable diagnostics. However, few competitions explicitly integrate educational QA, symbolic reasoning, and explainability within a unified framework. Table~\ref{tab:xai_comparison} provides a comparative overview of the XAI Challenge 2025 in relation to a selection of prominent AI competitions.

\begin{table}[ht]
\centering
\caption{Comparison of XAI Challenge 2025 with other notable AI competitions}
\label{tab:xai_comparison}
\small
\resizebox{\textwidth}{!}{%
\begin{tabular}{@{}L{4.8cm}L{3.5cm}M{1.2cm}M{1.7cm}M{1.5cm}L{3.1cm}@{}}
\toprule
\textbf{Competition} & \textbf{Field} & \textbf{QA-focused} & \textbf{Symbolic Reasoning} & \textbf{NL Explanation} & \textbf{Real-world Domain} \\
\midrule
EfficientQA \cite{pmlr-v133-min21a}  
 & Open-domain QA & \checkmark & \xmark & \xmark & General knowledge \\
Alexa Prize TaskBot\footnotemark[6]  
 & Task-based dialogue & \checkmark & \checkmark & \xmark & Household tasks \\
ARC\footnotemark[7] / Neuro-Symbolic ARC \cite{batorski2025nsa} 
 & Abstract reasoning & \xmark & \checkmark & \xmark & Synthetic puzzles \\
MIDRC XAI Challenge\footnotemark[8]  
 & Medical image diagnosis & \xmark & \xmark & \xmark & X-ray diagnostics \\
XAI Hackathon Pisa\footnotemark[10] 
 & XAI prototyping & \xmark & --- & --- & Mixed datasets \\
Explainable Fuzzy AI Challenge\footnotemark[11] 
 & Fuzzy logic systems & \xmark & \checkmark & \xmark & Agent control \\
xAI Challenge Austin\footnotemark[12] 
 & Coding + explanation & --- & \xmark & --- & General AI tasks \\
\rowcolor{Grey!10}
Our XAI Challenge 
 & Educational QA & \checkmark & \checkmark & \checkmark & Academic policy \\
\bottomrule
\end{tabular}%
}
\begin{flushleft}
\textbf{Legend:} \checkmark = Fully addressed; \xmark = Not addressed; --- = Partial or optional inclusion
\end{flushleft}
\end{table}

\footnotetext[10]{\url{http://xai-hackathon.isti.cnr.it/}}
\footnotetext[11]{\url{https://xfuzzycomp.github.io/XFC/}}
\footnotetext[12]{\url{https://finch-mauve-rk3e.squarespace.com/xai-atx}}

Our challenge introduces a unique combination of educational QA, symbolic reasoning, and logic-based natural language explanation. This combination is rarely emphasized simultaneously in other major competitions. Its requirement to use lightweight LLMs together with symbolic inference makes it particularly suitable for student-facing applications where transparency and trust are essential. In addition, the challenge's structure encourages interdisciplinary collaboration, rapid prototyping, and accountable system behavior. These qualities are often missing in leaderboard-driven competitions that focus narrowly on accuracy. Taken together, these attributes establish the XAI Challenge 2025 as a novel and practical benchmark for building explainable, trustworthy, and educationally aligned AI systems.

%% file: components/Objectives.tex
The XAI Challenge 2025 was motivated by the limitations of conventional LLM-based QA systems in educational settings. These systems often return concise answers with limited explanatory depth, making it difficult for users to trace errors in reasoning or verify the correctness of responses. This lack of transparency is particularly problematic in rule-based, high-stakes scenarios such as those involving university policies. To address this issue, the competition promoted hybrid approaches that combine the fluency of LLMs with the rigor of symbolic reasoning, with the goal of enhancing both interpretability and trustworthiness in educational QA systems. The primary objectives of the challenge are as follows.
\begin{itemize}
    \item \textbf{O1.} Encourage the development of QA systems that generate logic-grounded natural language explanations for policy-related queries.
    \item \textbf{O2.} Promote hybrid architectures that integrate symbolic inference with LLM-based generation.
    \item \textbf{O3.} Enhance the transparency and verifiability of automated responses in educational contexts.
    \item \textbf{O4.} Showcase real-world applications where explainability improves student comprehension and learning outcomes.
    \item \textbf{O5.} Recognize outstanding solutions through academic dissemination, including presentations at the TRNS-AI workshop and paper submissions to the \textit{4$^{\text{th}}$ Italian Conference on Big Data and Data Science} (ITADATA 2025).
\end{itemize}

%% file: components/Event_Structure.tex
The event was structured to foster rapid prototyping, interdisciplinary collaboration, and real-world impact. While inspired by the fast-paced spirit of traditional hackathons, the competition unfolded across multiple phases over a three-month period, allowing participants to iteratively refine their solutions. The complete timeline of the XAI Challenge 2025 is summarized in Table~\ref{tab:timeline}.

The challenge welcomed a diverse pool of participants, including high school and university students as well as early-career researchers with interests in XAI. Teams of up to six members could register between March 2 and April 25, 2025, with individuals without a team having the option to be matched into groups by the organizers. To support effective system development, a virtual kickoff workshop and dataset release were held on April 13. The main competition phase ran from April 14 to May 11, during which participants developed educational QA systems capable of answering university policy-related questions while generating logically grounded natural language explanations. Evaluation was conducted in two stages: (i) Phase 1 results were announced on May 12–13, followed by a brief model refinement period on May 14–15; and (ii) Phase 2 results were released on May 16–17, with final rankings determined on May 18. On June 1, a public test day was held, during which all systems were evaluated on a hidden test set. Each team also delivered a live presentation of their solution, followed by a Q\&A session with a panel of challenge chairs comprising renowned professors. Final rankings and awards were officially announced at the end of the day.

To extend the competition’s impact beyond its runtime, the top three teams were invited to present at the TRNS-AI Workshop (held as part of IJCNN 2025) on July 5 and to submit a full paper to the ITADATA Conference by June 30. These post-challenge activities were designed to encourage continued academic dissemination and foster sustained community engagement.

\begin{table}[ht]
\centering
\caption{Timeline of the XAI Challenge 2025}
\label{tab:timeline}
\small
\begin{tabular}{@{}ll@{}}
\toprule
\textbf{Date(s)} & \textbf{Event} \\
\midrule
March 2 -- April 25     & Team registration period \\
April 13               & Kickoff workshop and dataset release \\
April 14 -- May 11      & Main competition phase \\
May 12--13              & Phase 1 evaluation results \\
May 14--15              & Model refinement period \\
May 16--17              & Phase 2 evaluation results \\
May 18                 & Final ranking announcement \\
June 1                 & Public test day, solution presentations, and final result release \\
June 30                & Paper submission (Top 3 teams, ITADATA Conference) \\
July 5                 & Presentation at TRNS-AI Workshop (IJCNN 2025) \\
\bottomrule
\end{tabular}
\end{table}

%% file: components/Dataset.tex
\begin{figure}[!ht]
    \centering
\begin{prettyjson}
{
  "premises-NL": [
    "Every student enrolled in the course who completes at least 80
    "If a student attends all lectures, then they have a higher chance of passing the final exam.",
    "If a student attends a tutoring session or completes extra practice problems, they are more likely to improve their grades.",
    "No student who fails to submit their research paper passes the course.",
    "There exists a student who is on academic probation and later graduates with honors."
  ],
  "premises-FOL": [
    "ForAll(x, (Student(x) AND Completed80PctAssignments(x)) -> PassCourse(x))",
    "ForAll(x, AttendsAllLectures(x) -> HigherChancePassFinalExam(x))",
    "ForAll(x, (AttendsTutoringSession(x) OR CompletesExtraPractice(x)) -> MoreLikelyImproveGrades(x))",
    "ForAll(x, NOT SubmitsResearchPaper(x) -> NOT PassCourse(x))",
    "Exists(x, OnAcademicProbation(x) AND GraduatesWithHonors(x))"
  ],
  "questions": [
    "Which statement can be inferred?\nA. ...\nB. If a student attends a tutoring session and completes at least 80
    "Is this statement true?\nStatement: If a student attends all lectures but does not submit their research paper, they still cannot pass the course even though they have a higher chance of passing the final exam."
  ],
  "answers": [
    "B",
    "Yes"
  ],
  "idx": [
    [1, 3],   
    [2, 4]    
  ],
  "explanation": [
    "Premise 3. states that attending tutoring (or doing extra practice) increases the likelihood of grade improvement. Premise 1. says completing at least 80
    "Premise 2. says attending all lectures raises the chance of passing the final exam, but Premise 4. says any student who fails to submit the research paper cannot pass the course. Both conditions can hold simultaneously, so the statement is true."
  ]
}

\end{prettyjson}
    \caption{Example record from the XAI Challenge 2025 dataset. Each item includes natural and formal premises, natural language questions, indexed supporting evidence, and human-readable explanations.}
    \label{fig:example-record}
\end{figure}

To align with the educational goals of the challenge, the dataset was carefully constructed to reflect realistic, policy-based questions that students often face in academic settings. Each entry consists of a collection of premises presented in both natural language and \textit{First-Order Logic} (FOL), accompanied by one or more questions and their corresponding ground truth answers. The content covers a wide spectrum of university regulations, including course enrollment criteria, graduation requirements, credit thresholds, and exceptions for special circumstances. This design ensures that the dataset is both logically rigorous and pedagogically relevant. Questions are divided into three main categories:
\begin{itemize}
    \item \textbf{Yes/No/Uncertain:} Binary evaluations of compound logical conditions (e.g., academic eligibility or rule violations).
    \item \textbf{Multiple-choice:} Selecting the most logically entailed conclusion from a list of candidates.
    \item \textbf{Numerical:} Inferring specific quantities (e.g., credit totals or counts) from constraints embedded in the premises.
\end{itemize}

Figure~\ref{fig:example-record} illustrates a simple example from the dataset. Each record includes five core components: (i) premises expressed in both natural language (\texttt{premises-NL}) and formal logic (\texttt{premises-FOL}); (ii) one or more \texttt{questions} designed to test reasoning ability; (iii) ground truth \texttt{answers}; (iv) supporting premise indices (\texttt{idx}) that identify which statements justify each answer; and (v) a natural language \texttt{explanation} that outlines the reasoning steps in a transparent and verifiable manner. This structure encourages systems to generate answers based on explicit logical evidence rather than statistical approximation. Moreover, the data format is directly aligned with the challenge’s evaluation framework, which emphasizes not only answer accuracy but also the clarity and traceability of the accompanying explanations.

The dataset was built through a three-stage pipeline that combines symbolic reasoning, LLMs, and expert validation. In the first stage, a custom logic engine built on top of the Z3 Solver was used to generate logically consistent premises. The engine applied classical inference rules such as Modus Ponens, Hypothetical Syllogism, and De Morgan’s Theorem to construct a diverse set of original, derived, and unrelated statements. The full procedure is described in Algorithm~\ref{alg:premise-gen}, which outlines how premises were sampled, validated, and assembled into complete records. In the second stage, each validated FOL statement was translated into natural language using ChatGPT. These translations aimed to preserve the original logical meaning while rendering the statements in fluent, readable English. In the final stage, trained university students manually reviewed and refined the records to ensure clarity, factual accuracy, and contextual relevance.

\begin{algorithm}[ht]
\caption{Premise Generation for XAI Challenge Dataset}
\label{alg:premise-gen}
\SetKwInOut{Input}{Input}
\SetKwInOut{Output}{Output}

\Input{Number of steps $s$, number of chained premises $c$, number of derived premises $d$}
\Output{Dictionary \texttt{\{original, derived, unrelated\}} containing valid premises}

Initialize empty sets: \texttt{original}, \texttt{derived}, \texttt{unrelated}\;
Define logic variables (e.g., $P$, $Q$, $R$) and inference rules\;

\BlankLine
\textbf{Step 1: Generate original premises}\;
\For{$i \gets 1$ \KwTo $s$}{
  Generate a premise using a random inference rule\;
  Validate with Z3\;
  \If{valid and unique}{
    Add to \texttt{original}\;
  }
}

\BlankLine
\textbf{Step 2: Derive new premises}\;
\For{$i \gets 1$ \KwTo $d$}{
  Select two premises from \texttt{original} or \texttt{derived}\;
  Derive a new premise, e.g., using implication or conjunction\;
  Validate with Z3\;
  \If{valid and unique}{
    Add to \texttt{derived}\;
  }
}

\BlankLine
\textbf{Step 3: Add unrelated premises}\;
\For{$i \gets 1$ \KwTo $s - c$}{
  Generate an unrelated premise using a random rule\;
  Validate with Z3\;
  \If{valid and unique}{
    Add to \texttt{unrelated}\;
  }
}

\BlankLine
\Return \texttt{\{original, derived, unrelated\}}\;
\end{algorithm}

The finalized dataset consists of 481 training records and 50 test records. Each record combines various types of premises and question formats, designed to assess both factual comprehension and multi-step logical reasoning. Table~\ref{tab:dataset_stats} provides an overview of the dataset's composition, including statistics on premise length, question distribution, and reasoning depth.

\begin{table}[ht]
\centering
\caption{Dataset Statistics for the XAI Challenge 2025}
\label{tab:dataset_stats}
\small
\begin{tabular}{@{}lcc@{}}
\toprule
\textbf{Metric} & \textbf{Training Set} & \textbf{Test Set} \\
\midrule
Total Records & 481 & 50 \\
Average Premise Count per Record & 9.90 & 6.08 \\
Average Premise Length (Words) & 126.60 & 85.16 \\
Yes/No/Uncertain Records & 457 & 13 \\
Multiple-Choice Records & 403 & 21 \\
Numerical Records & 16 & 16 \\
Maximum Inference Steps & 20 & 6 \\
Maximum Premises per Record & 36 & 10 \\
\bottomrule
\end{tabular}
\end{table}

%% file: components/Rules.tex
To ensure transparency, fairness, and alignment with the goals of XAI, the challenge introduced a clear set of modeling and technical constraints. Participants were required to build educational QA systems that could answer university policy questions and generate natural language explanations grounded in specific evidence. Each system received input in the form of a JSON object with two fields:
\begin{itemize}
    \item \texttt{question}: a natural language question about academic policy,
    \item \texttt{premises}: a list of premises written in natural language.
\end{itemize}
The system was expected to return a JSON object with:
\begin{itemize}
    \item \texttt{answer}: the predicted answer (e.g., Yes, No, Uncertain, a number, or a multiple-choice letter),
    \item \texttt{idx}: indices of the premises that support the answer, using one-based indexing,
    \item \texttt{explanation}: a concise, human-readable justification derived from the cited premises.
\end{itemize}

To support fair comparison and prevent data leakage, participants were provided with only the training portion of the dataset. The test set was kept private and used exclusively for final evaluation during the competition. All models were required to operate strictly on the given set of premises. External retrieval or lookup mechanisms were not allowed. To encourage transparency and discourage black box behavior, participants were recommended to use interpretable reasoning methods. These included symbolic solvers (such as Z3), lightweight open-source language models, or hybrid combinations. Regardless of approach, each system had to identify which premises were used and explain the reasoning process in a way that was understandable to non-expert users, especially students. In addition to modeling guidelines, each submission had to satisfy technical requirements to ensure reproducibility, robustness, and fairness during evaluation. Each system was deployed as an HTTP API and automatically tested on both private and public test cases. Table~\ref{tab:rules-constraints} outlines the main rules applied to all submissions.

\begin{table}[ht]
\caption{Summary of rules and constraints in the XAI Challenge 2025}
\label{tab:rules-constraints}
\centering
\small
\begin{tabular}{@{}p{3.2cm}p{11cm}@{}}
\toprule
\textbf{Aspect} & \textbf{Constraint} \\
\midrule
Model transparency & Only open-source models with fewer than 8 billion parameters were allowed without penalty. Submissions based entirely on proprietary models (e.g., GPT\footnotemark[13], DeepSeek\footnotemark[14]) were ranked lower to encourage reproducibility. \\
Reasoning method & Systems were required to generate natural language justifications grounded in specific premises. The use of symbolic solvers, lightweight language models, or hybrid systems was encouraged. \\
External data & Any external data used for training, fine-tuning, or augmentation had to be fully disclosed, including the source and intended use. Non-disclosure resulted in disqualification. \\
API protocol & Each system had to expose an API accepting \texttt{POST} requests with a JSON input containing a question and list of premises. \\
API output & The response had to include: (i) the predicted answer, (ii) one-based indices of the supporting premises, and (iii) a concise, human-readable explanation. \\
Lookup tables & Hardcoded or static responses were prohibited. All outputs had to be computed dynamically. \\
Rate limit & APIs were limited to 10 requests per second, with a maximum processing time of 60 seconds per request. \\
Availability & APIs that were offline for more than 30 consecutive minutes or failed more than 10\% of test queries were disqualified. \\
\bottomrule
\end{tabular}
\end{table}

\footnotetext[13]{\url{https://openai.com/index/openai-api/}}  
\footnotetext[14]{\url{https://platform.deepseek.com/}}

%% file: components/Evaluation.tex
 As outlined in Table~\ref{tab:timeline}, our challenge employed a multi-phase evaluation framework to assess participants' systems through hosted API endpoints, using both private and public datasets. The evaluation focused not only on answer accuracy but also on reasoning transparency and explanatory clarity. Each system was assessed along three primary dimensions:
\begin{itemize}
    \item \textbf{Correctness of Answers ($P_1$)} — measured using \textit{Exact Match} (EM) between the system's output and the ground-truth \texttt{answer} field.

    \item \textbf{Relevance of Premises ($P_2$)} — evaluated via EM with the ground-truth \texttt{idx} field (one-based indexing), assessing whether the system selects the minimal correct subset of premises that support the answer.

    \item \textbf{Explainability ($P_3$)} — evaluates the clarity and logical coherence of the generated natural language \texttt{explanation}, which must be concise, faithful to the selected premises, and comprehensible to human users. In the final round, $P_3$ was manually scored by the panel of professors based on a rubric.
\end{itemize}

Each of $P_1$, $P_2$, and $P_3$ is computed per instance and normalized to the range $[0, 1]$. Since each question in the dataset has exactly one correct answer and one minimal supporting set of premises (generated via logic-based templates and validated with symbolic solvers), exact match is a reliable metric for $P_1$ and $P_2$. To ensure logical consistency, we enforce the constraint defined in Equation~\ref{eq:consistency}.
\begin{equation}
\label{eq:consistency}
P_1 \cdot P_2 = 0 \Rightarrow \text{score} = 0
\end{equation}

\noindent\textbf{Selection Round.}\quad This round consisted of two sub-phases. In each phase, systems were evaluated on the same set of $n = 50$ private test cases using the same scoring scheme. For each instance, the score $s_i$ was computed as shown in Equation~\ref{eq:s_inst}.
\begin{equation}
\label{eq:s_inst}
s_i = 0.5 \cdot P_1 + 0.5 \cdot P_2
\end{equation}

The total score for each phase, denoted $S^{(1)}$ and $S^{(2)}$, was computed by summing over all test cases, as defined in Equation~\ref{eq:s1_s2}.
\begin{equation}
\label{eq:s1_s2}
S^{(k)} = \sum_{i=1}^{50} s_i^{(k)} \quad \text{for } k \in \{1, 2\}
\end{equation}

After Phase 1, participants received feedback and had the opportunity to refine their systems before re-submission in Phase 2. To further encourage a deep understanding of the dataset and its logical structure, we also introduced a dataset feedback incentive. Specifically, a bonus score $S_{\text{bonus}}$ was awarded based on the number and quality of valid issues (e.g., annotation errors, ambiguous premises) reported by each team. Although the dataset underwent rigorous validation, such community-driven review helped further improve its quality before public release. The final selection score $S_1$ used to determine advancement was computed as shown in Equation \ref{eq:s1_final}. 
\begin{equation}
\label{eq:s1_final}
S_1 = 0.6 \cdot \left(0.7 \cdot S^{(1)} + 0.3 \cdot S_{\text{bonus}}^{(1)}\right) + 0.4 \cdot \left(0.9\cdot S^{(2)} + 0.1\cdot S_{\text{bonus}}^{(2)}\right) 
\end{equation}

The five teams with the highest $S_1$ scores were invited to the final round.

\medskip
\noindent\textbf{Final Round.}\quad In this round, each system was evaluated on $n = 5$ public test cases and a live presentation session. For each test case, the instance-level score $s_i$ was computed as shown in Equation~\ref{eq:s2_inst}.
\begin{equation}
\label{eq:s2_inst}
s_i = 0.5 \cdot P_1 + 0.3 \cdot P_2 + 0.2 \cdot P_3
\end{equation}

The total model score $S_2$ was computed by summing over all test cases, as shown in Equation~\ref{eq:s2}.
\begin{equation}
\label{eq:s2}
S_2 = \sum_{i=1}^{5} s_i
\end{equation}

Each team also gave a live presentation, which was evaluated in two parts: a 7-minute technical presentation and a Q\&A session with a panel of professors. Each part was independently scored by the judges using a 5-point Likert scale (1 = very poor, 5 = excellent), based on criteria such as clarity, content depth, delivery quality, and responsiveness. Let $R_{\text{pres}}$ and $R_{\text{Q\&A}}$ denote the average rubric scores for the presentation and Q\&A portions, respectively. The final presentation score $S_3$ was computed as shown in Equation~\ref{eq:presentation_raw}.
\begin{equation}
\label{eq:presentation_raw}
S_3 = \frac{R_{\text{pres}} + R_{\text{Q\&A}}}{10}
\end{equation}

The overall final score $S$ used for ranking was computed as the average of model and presentation components, defined in Equation~\ref{eq:final}.
\begin{equation}
\label{eq:final}
S = 0.5 \cdot S_2 + 0.5 \cdot S_3
\end{equation}

This composite score $S$ determined the final team rankings and award decisions.

%% file: components/Participants.tex
The XAI Challenge 2025 attracted a diverse cohort of 107 participants, organized into 28 teams. Team sizes ranged from individual participants to groups of up to six members, enabling a variety of collaboration formats. This flexible structure allowed contributions from individuals across a broad spectrum of backgrounds and experience levels, including undergraduate students, graduate students, and early-career researchers. While most participants were based in Vietnam, the host country, and India, the challenge also welcomed teams from other countries. This international participation underscored the growing global relevance of XAI in education. 

%% file: components/Results.tex
Table~\ref{tab:selection-summary} reports the performance of the top five finalists across both sub-phases of the Selection Round. Each system was evaluated on the same set of 50 private test cases per phase, following the scoring procedure described in Section~\ref{sec:evaluation}. 

\begin{table}[ht]
\centering
\caption{Performance of top five finalists in the Selection Round (anonymized)}
\label{tab:selection-summary}
\small
\begin{tabular}{lcccc}
\toprule
\textbf{Team} & \textbf{Phase 1 Score} & \textbf{Phase 2 Score} & \textbf{Final Selection Score} & \textbf{Final Round Rank} \\
\midrule
Team A & 19.90 & 21.00 & 20.34 & 2 \\
Team B & 19.10 & 22.05 & 20.28 & 1 \\
Team C & 13.85 & 25.80 & 18.63 & 4 \\
Team D & 17.60 & 18.45 & 17.94 & 5 \\
Team E & 18.45 & 16.65 & 17.73 & 3 \\
\bottomrule
\end{tabular}
\end{table}

\medskip
\noindent\textbf{Modest Scores Reflect the Task’s Inherent Difficulty.} \quad
Although each evaluation phase included only 50 test cases, the absolute scores across all teams remained modest. The highest final selection score barely surpassed 20, representing less than 50\% of the maximum possible. This outcome highlights the intrinsic complexity of the task, which required systems not only to produce correct answers, but also to identify a minimal set of supporting premises and generate logically sound, human-understandable explanations. The results underscore the broader challenge of developing AI systems capable of faithful and interpretable reasoning over real-world educational policies.

\medskip
\noindent\textbf{Two-Phase Evaluation Facilitated Iterative Improvement.} \quad
The two-phase structure gave teams an opportunity to refine their systems under tight time constraints. Notably, Team C demonstrated substantial improvement, raising its Phase 2 score by nearly 86\% over Phase 1. This suggests that meaningful progress was achievable through targeted model updates and deeper engagement with the dataset. In contrast, Team E saw a decline in performance, showing that not all adjustments led to better results and highlighting the risk of overfitting or ineffective tuning during rapid iteration.

\medskip
\noindent\textbf{Selection Scores Were Not Always Predictive of Final Rankings.} \quad
There was no strict correlation between selection scores and final round outcomes. For instance, Team A, which led the Selection Round, was ultimately overtaken by Team B in the Final Round. Meanwhile, Team E preserved its selection rank but narrowed the gap with top teams. These shifts reflect the multifaceted nature of the Final Round, where systems were evaluated not only on public test cases but also through live presentations assessed by a panel of experts. As a result, final rankings depended not just on model output, but also on a team’s ability to explain its design choices, reasoning strategies, and approach to explainability. This underscores the comprehensive and demanding nature of the challenge.

%% file: components/Solution.tex
This section highlights several representative systems developed during the XAI Challenge 2025. While differing in architecture and methodology, these systems shared the common goal of producing accurate answers accompanied by logically grounded and interpretable explanations. We summarize four notable design paradigms that illustrate the diversity of strategies explored by the finalists.

\medskip
\noindent\textbf{Multi-Agent Systems with Symbolic Reasoning.} \quad  
One top-performing team implemented a modular multi-agent system that combined several lightweight open-source LLMs with symbolic reasoning via the Z3 theorem prover. The system divided the QA pipeline into specialized components: one agent parsed natural language premises, another applied logical inference using Z3, and a third synthesized structured explanations. Intermediate outputs were passed between agents to ensure modularity and traceability. This approach proved especially effective in handling complex queries involving conditional rules and policy exceptions, contributing to the team’s first-place result.

\medskip
\noindent\textbf{Prompt-Based Learning with Task-Specific Templates.} \quad  
Several teams adopted prompt-based learning, using carefully designed templates to guide lightweight LLMs in extracting relevant premises and generating step-by-step explanations. Prompts were tailored to specific question types, such as Yes/No/Uncertain, multiple-choice, or numerical answers. Some systems further employed Chain-of-Thought (CoT) \cite{NEURIPS2022_9d560961} prompting to encourage intermediate reasoning steps. This strategy offered a lightweight and interpretable solution but remained constrained by the inherent opacity and prompt sensitivity of black-box models.

\medskip
\noindent\textbf{Rule Retrieval with Symbolic Inference.} \quad  
Another approach focused on rule retrieval and symbolic logic. One team built a structured rulebase of educational policies in Python and used keyword or semantic matching to identify candidate premises. These were then passed to the Z3 solver for formal inference. Explanations were generated by mapping the solver’s logical steps to human-readable justifications. This method performed well on regulation-heavy queries but showed limitations when dealing with ambiguous or loosely structured inputs.

\medskip
\noindent\textbf{Multi-Task Fine-Tuning with a Mixture-of-Experts Architecture.} \quad  
A learning-based system fine-tuned multiple lightweight LLMs on synthetic supervision for three distinct tasks: answer generation, premise selection, and explanation construction. These tasks were routed through a \textit{Mixture-of-Experts} (MoE) \cite{jiang2024mixtral} architecture, where each expert model specialized in one task and was activated based on input type. While this approach benefited from task-specific learning and synthetic data control, it lacked the interpretability of symbolic systems and remained reliant on black-box inference.

\medskip
These approaches illustrate a range of trade-offs between transparency, flexibility, and performance. Systems that integrated symbolic reasoning provided clear, verifiable explanations, while those based on language models offered adaptability and linguistic fluency. The challenge encouraged exploration across this design space, yielding valuable insights into the development of XAI systems for education and policy domains.

%% file: components/Conclusion.tex
The XAI Challenge 2025 highlighted the feasibility and significance of developing transparent QA systems for educational contexts. By imposing constraints on model size and requiring verifiable reasoning, the competition challenged participants to design solutions that balanced accuracy with interpretability and trustworthiness.

The event drew a variety of approaches, including multi-agent architectures, prompt-based learning pipelines, rule-driven retrieval systems, and multi-task fine-tuning strategies. Despite their differences, these systems shared a common objective: to integrate natural language understanding with logical inference in order to answer policy-related queries with clarity and justification. A key design constraint was explainability, enforced through open-source implementation and the requirement to output explicit reasoning chains. As a result, models were evaluated not only on the correctness of their answers but also on their ability to select supporting premises and articulate coherent explanations.

Taken together, the outcomes of the challenge provide meaningful insights into the design of XAI systems in high-stakes settings. The event demonstrates that \textit{bridging LLMs and symbolic reasoning is not only possible, but can yield practical, interpretable solutions to real-world tasks such as educational QA}. This serves as a foundation for future research at the intersection of language understanding, reasoning, and explainability.